\documentclass[10pt,twocolumn,letterpaper]{article}

\usepackage{cvpr}              

\usepackage{graphicx}
\usepackage{amsmath}
\usepackage{amssymb}
\usepackage{booktabs}
\usepackage{multirow}
\usepackage{mathtools} 
\usepackage{multicol}
\usepackage{pifont}
\usepackage{color, colortbl}
\newcommand{\xmark}{\ding{55}}
\definecolor{Gray}{gray}{0.9}
\usepackage{enumitem}

\definecolor{cvprblue}{rgb}{0.21,0.49,0.74}
\usepackage[pagebackref,breaklinks,colorlinks,allcolors=cvprblue]{hyperref}

\title{ReWind: Understanding Long Videos with Instructed Learnable Memory}

\author{
Anxhelo Diko$^{1}$\thanks{Equal Contribution} \thanks{Work done while at Huawei} \quad Tinghuai Wang$^{2}$\footnotemark[1] \quad Wassim Swaileh$^{2}$ \quad Shiyan Sun$^2$ \quad Ioannis Patras$^{2}$ \\
$^1${\small La Sapienza University of Roma}
$^2${\small Huawei Helsinki Research Center} \\
{\tt\small diko@di.uniroma1.it \{tinghuaiwang,shiyansun,wassim.swaileh,ioannis.patras\}@huawei.com}
}
\begin{document}

\maketitle

\renewcommand{\baselinestretch}{0.9}

\begin{abstract}

Vision-Language Models (VLMs) are crucial for applications requiring integrated understanding textual and visual information. 
However, existing VLMs struggle with long videos due to computational inefficiency, memory limitations, and difficulties in maintaining coherent understanding across extended sequences. To address these challenges, we introduce ReWind, a novel memory-based VLM designed for efficient long video understanding while preserving temporal fidelity. ReWind operates in a two-stage framework. In the first stage, ReWind maintains a dynamic learnable memory module with a novel \textbf{read-perceive-write} cycle that stores and updates instruction-relevant visual information as the video unfolds.
This module utilizes learnable queries and cross-attentions between memory contents and the input stream, ensuring low memory requirements by scaling linearly with the number of tokens. In the second stage, we propose an adaptive frame selection mechanism guided by the memory content to identify instruction-relevant key moments. It enriches the memory representations with detailed spatial information by selecting a few high-resolution frames, which are then combined with the memory contents and fed into a Large Language Model (LLM) to generate the final answer. We empirically demonstrate ReWind's superior performance in visual question answering (VQA) and temporal grounding tasks, surpassing previous methods on long video benchmarks. Notably, ReWind achieves a +13\% score gain and a +12\% accuracy improvement on the MovieChat-1K VQA dataset and an +8\% mIoU increase on Charades-STA for temporal grounding.

\end{abstract}


\section{Introduction}
\label{s:intro}

Large Language Models (LLMs) \cite{touvron2023llama,team2024gemma} have demonstrated remarkable capabilities at human language processing \cite{vaswani2017attention,bommasani2021opportunities}. However, these models are limited to text-based inputs and, therefore, oblivious to real-world, multi-sensory information. To address this limitation, researchers are actively developing Multimodal LLMs (MLLMs) capable of processing signals from multiple and diverse modalities \cite{li2023blip,li2023llamavid,maaz2024ChatGPT}, including images, video, and audio. This emerging field holds immense potential for applications such as visual question answering (VQA), real-time interfaces for autonomous agents, and generating detailed scene descriptions for the visually impaired.

\begin{figure}[!t]
\includegraphics[width=\linewidth]{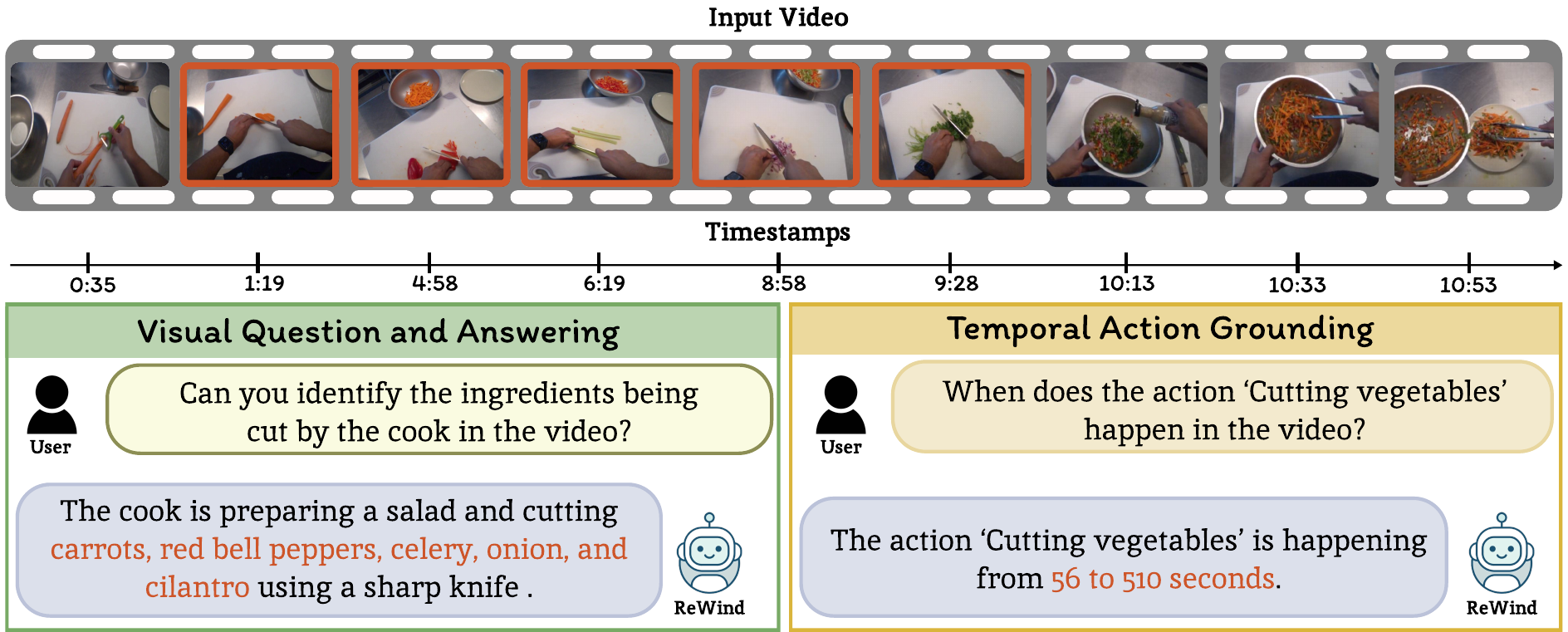}
 \caption{ReWind is a memory-based VLM framework designed for long video understanding (10+ minutes), specialized in VQA and temporal grounding. The highlighted frames are selected from ReWind's dynamic frame selection mechanism.}
 \label{fig:teaser}
  \vspace{-5mm}
\end{figure}

Recent research in MLLMs has predominantly concentrated on Vison-Language Models (VLMs) \cite{zhang2023videollama,zhang2024llamaadapter,huang2024vtimellm}, which typically combine pre-trained LLMs with visual encoders that encode and feed to them visual information. 
However, existing VLMs face two major challenges in processing long videos. First, their self-attention mechanisms require substantial memory that scales quadratically with the number of tokens, making long video processing computationally intensive. Second, these models struggle to effectively model temporal dependencies over extended sequences.
To address these challenges, recent efforts have proposed using memory modules to enhance the capability of VLMs \cite{song2024moviechat,he2024ma}. However, current memory modules often serve as storage units and lack the ability to discern and retain information pertinent to the task or user instructions. Moreover, these models tend to compress temporal information heavily \cite{song2024moviechat}, sacrificing the fidelity of the temporal dynamics and overlooking critical details in the video's narrative. Additionally, current models rely on fixed dense spatial representations per frame \cite{he2024ma,song2024moviechat}, increasing memory requirements: by treating all frames equally, they store unnecessary details for non-essential moments, increasing memory demands and limiting the model's ability to focus on critical events for accurate video comprehension.

To address these long video challenges, we introduce \textit{ReWind}, a novel memory-based framework that operates in two stages, advancing the state of the art with key innovations. In the first stage (Stage-1 in Fig. 2), through a learnable memory module, Rewind enables instruction-guided feature encoding and storage into a memory bank of coherent temporal information. At its core, ReWind features a novel \textit{read-perceive-write} cycle: First (Read Cross Attention), a \textit{read} operation looks at historical context from memory and produces fixed-size read queries. Then those are used as queries in a \textit{perceiver} unit (Perceiver Block) that processes tokens from the encoder of incoming frames. Unlike previous Q-Former approaches (e.g.,\cite{zhang2023videollama}) that compress information at clip-level, our novel design allows memory-informed processing of the incoming frames and preserves temporal fidelity. Finally, the perceiver's representations of the input tokens flow into the \textit{write} operation (Write Cross Attention), where learnable write queries distill and filter information through. The resulting compact representations are then stored in memory, enabling ReWind to progressively build coherent temporal representations while avoiding the compression issues present in previous works \cite{song2024moviechat,he2024ma,zhang2023videollama}. Crucially, in this stage, we avoid cross-attention between the memory and the video stream, as well as self-attention within the stream tokens with high computational demand. In the second stage (Stage-2 in Fig. 2), ReWind 'rewinds' the video stream and dynamically selects frames by a selection mechanism guided by the memory contents and the user instructions. The selection mechanism operates on high spatial resolution tokens from the input stream so that after selection, tokens from both the memory bank and the dense selection outputs are fed into an LLM that generates the response. By contrast to previous works that maintain fixed-size dense representations for each frame \cite{song2024moviechat,zhang2023videollama}, this selection strategy incorporates detailed spatial information only for relevant key events, resulting in reduced memory requirements.

In our extensive evaluations, ReWind demonstrates superior performance compared to previous state-of-the-art methods across both long and short-term video question answering \cite{song2024moviechat,chen2011msvd,xu2016msr,2023videochat} and temporal grounding video benchmarks \cite{gao2017tall,caba2015activitynet}, validating the effectiveness of our approach. Additionally, detailed ablation motivates our design choices. In summary, the main contributions of this work are threefold:
\begin{itemize}
    \item ReWind, a novel memory-based vision-language model that enables efficient understanding of long videos while maintaining temporal fidelity.
    \item A learnable memory module with an innovative read-perceive-write cycle that enables instruction-guided feature encoding and robust temporal representation construction.
    \item An adaptive frame selection mechanism that identifies instruction-relevant key moments and enriches memory representations with detailed spatial information for comprehensive video understanding.
\end{itemize}

 
%
    
    

\section{Related Works}
\label{s:related_work}

\subsection{Short Video Understanding}\label{ss:svu}
Recent VLMs have explored various architectural approaches for video understanding.  Dual-stream architectures, exemplified by Video-LLaMA \cite{zhang2023videollama} and VideoChat \cite{2023videochat}, process different modalities separately.  The former processes both audio and visual information separately using Q-Formers \cite{zhou2022qformer}. The latter processes video using specialized embedding models and a perception toolkit for mixed modalities. In contrast, single-stream approaches like Video-ChatGPT \cite{maaz2024ChatGPT} employ spatiotemporal pooling to capture the overall video context. Video-LLaVA \cite{lin2023llava} utilizes a LanguageBind \cite{zhu2023languagebind} module to map multimodal inputs into a shared space. Mirasol3B \cite{piergiovanni2024mirasol3b} proposes a decoder-only model adapted to handle multimodal input, representing them in disentangled spaces. ChatUniVi \cite{jin2024chat} takes a unique approach by introducing a unified visual representation through dynamic visual tokens for both images and videos.

\begin{figure*}[!ht]
\includegraphics[width=\textwidth]{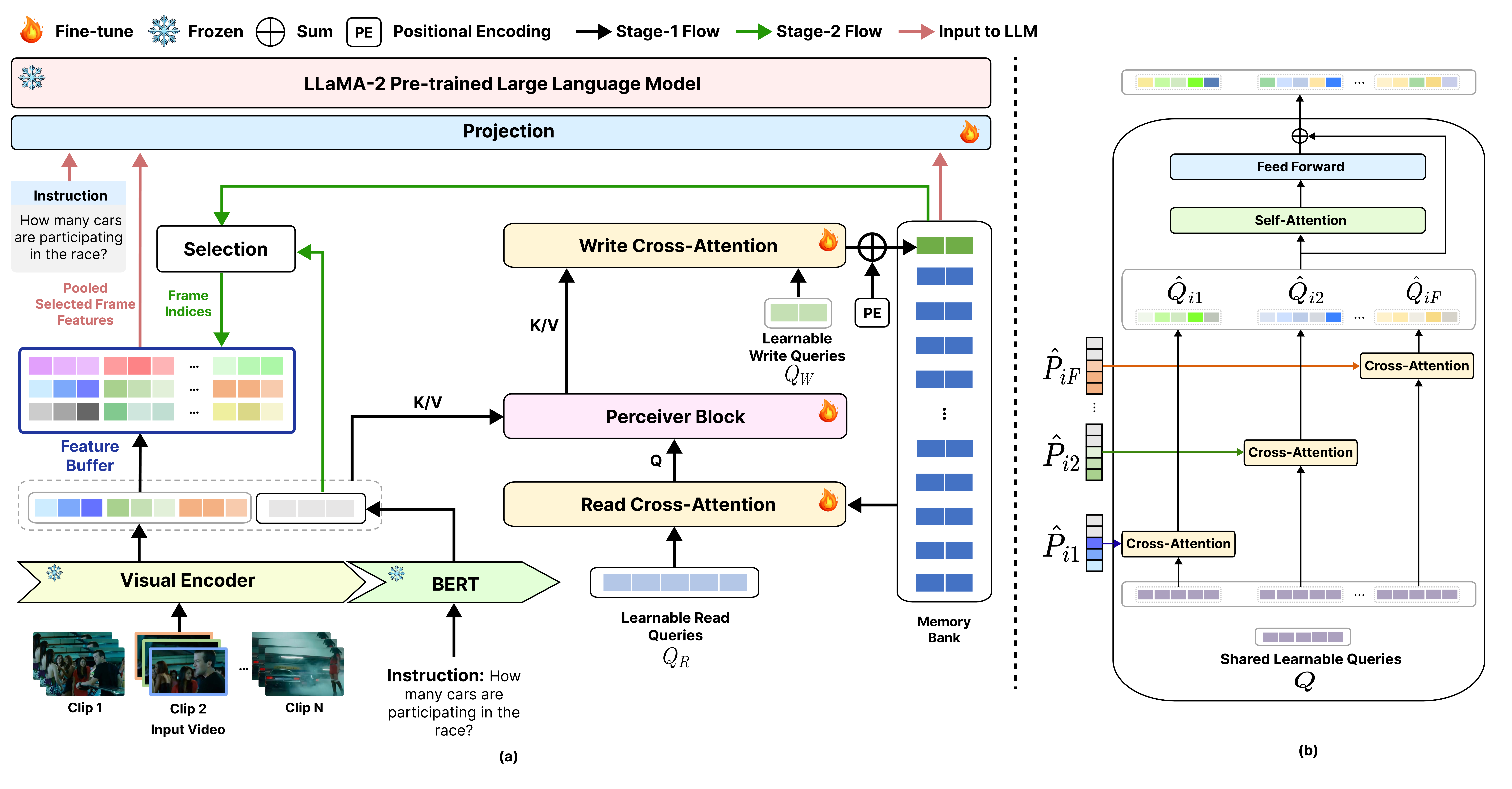}
 \caption{ReWind's VLM architecture for long video processing is illustrated in (a). It employs a two-stage processing scheme. In Stage 1 (black arrows), ReWind sequentially processes each video sub-clip using a visual encoder and a text-conditioned perceiver layer supported by a learnable memory module. This module performs read-and-write operations to ensure efficient information storage and maintain temporal coherence in a novel \textit{read-perceive-write} cycle. In Stage 2 (green arrows), ReWind utilizes a dynamic frame selection (DFS) mechanism to incorporate detailed spatial information for key moments. Finally (red arrow), the memory content, selected frames, and user instruction are combined to form the input for the language model. In (b), the perceiver layer with learnable queries and text-conditioned visual features for instruction-guided encoding.}
 \label{fig:arch}
  \vspace{-5mm}
\end{figure*}

\subsection{Long Video Understanding}

Recent works have proposed diverse solutions to address the challenges pertinent to long video understanding. Memory-based approaches include MovieChat \cite{song2024moviechat}, which employs a dual memory module with a FIFO queue for short-term memory and a consolidation module for long-term memory, and MA-LMM \cite{he2024ma}, which introduces a hierarchical memory module. TimeChat \cite{ren2024timechat} incorporates timestamps and transcribed speech for time-aware encoding. However, these approaches \cite{song2024moviechat,he2024ma,ren2024timechat} significantly compress temporal information, compromising the understanding of event dynamics. Alternative approaches focus on efficient frame representation. LLaMA-VID \cite{li2023llamavid} efficiently represents each frame with only two tokens. VTimeLLM \cite{huang2024vtimellm} introduces temporal-focused training and uses only the class tokens as frame representations. Yet, both VTimeLLM and LLaMA-VID process frames in isolation, failing to capture coherent temporal representations.

Unlike previous works that either significantly compress temporal information \cite{song2024moviechat,he2024ma,ren2024timechat} or process frames in isolation \cite{li2023llamavid}, ReWind distinguishes itself by proposing a novel memory-based architecture with a read-perceive-write cycle that selectively stores instruction-relevant visual information while enabling efficient processing of long videos and maintaining temporal fidelity. As opposed to approaches that maintain fixed dense representations \cite{song2024moviechat,he2024ma}, ReWind employs an adaptive frame selection mechanism that enriches memory representations with detailed spatial information only for instruction-relevant key moments.

\section{Method}
\label{s:method}

ReWind enables efficient long-video understanding through a novel memory-based architecture that maintains temporal fidelity while selectively storing instruction-relevant information. As shown in Fig. \ref{fig:arch} (a), the architecture implements this through two-stage processing. Stage-1, namely \textit{read-perceive-write} cycle, comprises: (1) a vision encoder, (2) a text encoder for instruction processing, (3) a instruction-aware perceiver that bridges visual features and LLM understanding, and (4) a memory module with learnable read and write operations for efficient information storage. Stage-2, the \textit{Selection}, comprises a dynamic frame selection (DFS) mechanism that enriches memory representations with detailed spatial information for key moments. These two stages work in concert to enable the LLM to generate responses based on both the instruction and video content. We explain Stage-1 components in Sections \ref{ss:vfe} and \ref{ss:memory}, and the DFS in Section \ref{ss:dfs}. Finally, we detail the LLM input formation and the training strategy in Sections \ref{ss:llm} and\ref{ss:training}.

\subsection{Visual Feature Extraction} 
\label{ss:vfe}
To process long videos under GPU memory constraints, ReWind divides input video $V$ containing $T$ frames into $N$ sub-clips $S = {s_1, s_2, \dots, s_N}$, each with $F$ frames ($N = T/F$). For each frame $f_{ij}$ in sub-clip $s_i$, a pre-trained ViT-G/14 encoder from EVA-CLIP \cite{fang2023eva} extracts visual features as a sequence of tokens $P_{ij}$.

\subsection{Instructed Memory Architecture}
\label{ss:memory}
\begin{figure}[!t]
\includegraphics[width=\linewidth]{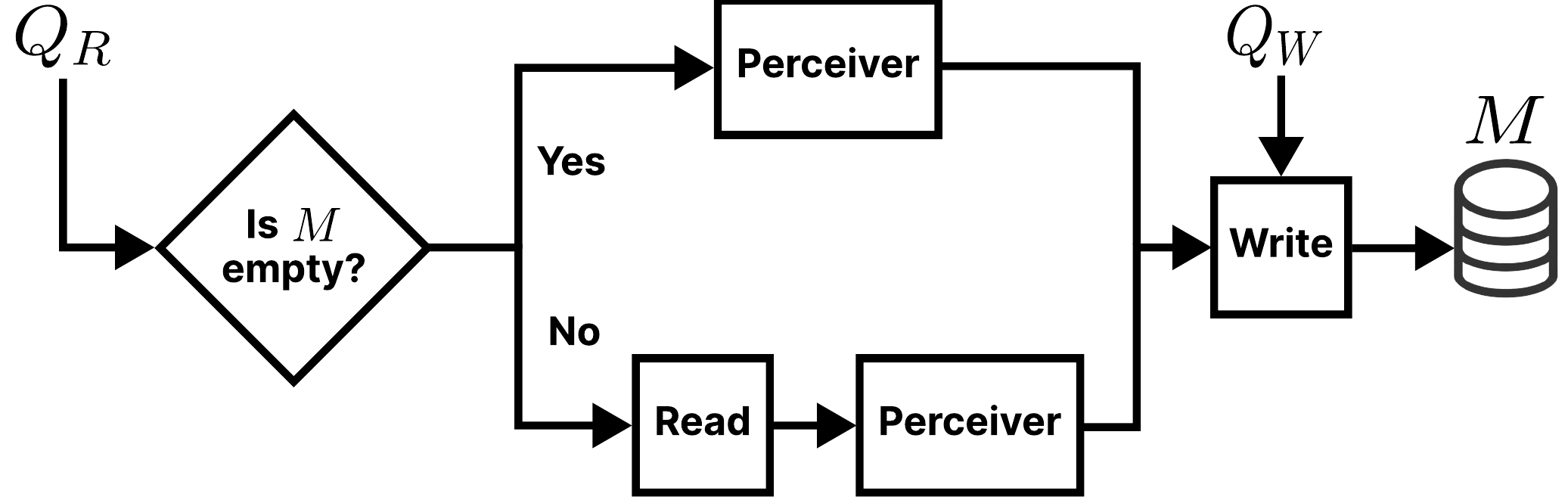}
 \caption{Rewind's \textit{read-perceive-write} simplified workflow.}
 \label{fig:memory_workflow}
  \vspace{-5mm}
\end{figure}

At the core of ReWind lies its novel read-perceive-write cycle that enables progressive video understanding while maintaining temporal fidelity. This cycle orchestrates the interaction between a long-term memory bank for storing distilled video representations, an instruction-aware temporal perceiver for temporal representation construction, and learnable read-write functions for memory interaction, as illustrated in Fig. \ref{fig:memory_workflow}.

To effectively process long videos, ReWind's memory module selectively stores instruction-relevant information from incoming frames while enabling progressive information accumulation. The module centers on a long-term memory bank $M$ and learnable read-write functions that bridge memory content with perceiver features. The read operation, using learnable queries $Q_R$, first retrieves historical context from $M$. These read queries then initialize the perceiver's queries for instruction-guided visual feature extraction from ViT outputs. Finally, learnable write queries $Q_W$ distill the perceiver's output through cross-attention for efficient storage in $M$. Additionally, original visual features are preserved in a feature buffer for potential detailed spatial analysis. This tight integration between memory operations and the perceiver ensures temporally coherent representations while maintaining computational efficiency.

\subsubsection{Read Operation}
\label{sss:reading}

The read operation aims to facilitate dynamic, context-aware feature extraction. This interface enables continuous interaction between the feature extraction process and the evolving memory content in $M$.  Specifically, as the memory gets populated with information from previously processed video segments, the read interface uses a fixed number $N_R$ (i.e., 32) of read queries $Q_R$ to actively retrieve relevant context through a cross-attention mechanism between them and the contents of $M$ as depicted in Fig. \ref{fig:arch} (a). This retrieval process enables the feature extraction pipeline to remain informed by the most recent knowledge stored in the memory.
These context-enriched read queries then guide the perceiver's processing of incoming frames, ensuring that feature extraction maintains awareness of previously stored temporal information.

\subsubsection{Perceive Operation}
\label{sss:perciever}
The perceive operation, performed by a perceiver block, bridges visual features and the LLM's understanding through instruction-aware temporal modeling. As illustrated in Figure \ref{fig:arch} (b), the design of perceiver allows for effective integration of instruction-guided features with historical context. As such, it utilizes a set of $N_Q$  learnable queries, $Q$, to project $P_{ij}$ into a latent space that LLM can understand. These learnable queries guide the extraction of relevant information from the visual features.

A crucial aspect of ReWind's design is the synergistic relationship between the perceiver block and the memory module. The learnable $Q$ in the perceiver block share the same weights and are initialized with the current content of the read queries $Q_R$ obtained by the cross attention between $Q_R$ and the contents of $M$ (note that this implies $N_Q = N_R$). This creates a continuous pipeline, allowing the feature extraction process to dynamically interact with the memory and access relevant context. To further enhance this process, the perceiver block incorporates the textual embedding of the user instruction, denoted as $I$. This embedding is obtained by encoding the input text query using a pre-trained BERT encoder. $I$ is then appended to the visual features $P_{ij}$ to form extended representations, denoted as $\hat{P_{ij}}$. As depicted in Fig. \ref{fig:arch} (b), the perceiver block employs a cross-attention mechanism between $Q$ and the combined visual-textual features $\hat{P_{ij}}$. This cross-attention allows the model to selectively attend to the most relevant aspects of the visual information conditioned on the user instruction. The output of this mechanism is a set of refined frame-level representations, denoted as $\hat{Q}_{ij}$.

Since $Q$ shares weights and content with updated read queries $Q_R$, the feature extraction process is also conditioned by the memory content. Specifically, as denoted in Fig. \ref{fig:arch} (a), $Q_R$ are always updated with the latest memory content before being used by the perceiver, assuming $M$ has content in it. This ensures a progressive construction of robust and temporally informed representations of the video content at the frame level.
Finally, to capture temporal relationships within the clip $s_i$, the perceiver performs self-attention on the temporal dimension of the refined representations $\hat{Q}_{ij}$ of consecutive frames. This temporal attention allows the model to understand how events unfold inside the clip. Note that unlike previous video Q-formers \cite{zhang2023videollama,song2024moviechat} that produce clip-level representations, our perceiver processes each frame individually and then performs temporal attention. This approach preserves temporal fidelity and enables a more nuanced understanding of event dynamics while keeping a robust representation of each frame.

\subsubsection{Write Operation}
\label{sss:write}
The write operation efficiently distills and stores the perceiver's frame-level output $\hat{Q}_{ij}$ in memory. While these outputs capture rich spatial and contextual information, their sheer number of queries impedes processing long videos.
To address this, ReWind's learnable writing mechanism compresses the visual information into a more efficient representation. This mechanism utilizes a set of learnable write queries, $Q_W$, to distill the scene information into a much smaller number of tokens (e.g., 2 tokens per frame). Specifically, ReWind employs cross-attention between $Q_W$ and $\hat{Q}_{ij}$ to generate compact per-frame representations $\hat{Q}^W_{ij}$. These representations are then stored in the memory bank $M$ in temporal order, enabling the progressive construction of temporally coherent video representations. Additionally, the original visual features $P_{ij}$ for each frame are stored in a separate feature buffer, preserving the detailed spatial information for later use (see Section \ref{ss:dfs}). This feature buffer is a simple storage container and does not impact computational resources.

\subsection{Dynamic Frame Selection}
\label{ss:dfs}
While $M$ efficiently stores a compressed video representation, certain instructions demand high spatial resolution at specific moments. ReWind addresses this through a Dynamic Frame Selection (DFS) mechanism that identifies instruction-relevant key frames using memory contents $M$ and instruction encoding $I$. This two-stage selection process, comprising instruction-based selection and clustering, occurs during \textit{Stage-2} of ReWind after full video processing and information storage in memory. 

\noindent\textbf{Instruction based selection.} The first stage prioritizes frames based on their relevance to the user's instruction by leveraging $I$, and contents of $M$. Given frame representations $\{m_t\}_{t=1}^{T} \in M$ and averaged instruction encoding $\overline{I}$, we compute the attention matrix between $\overline{I}$ and the contents of $M$. The top $L$ frames with the highest response scores to the instruction, denoted as $Z=\{z_l\}_{l=1}^L$, are then selected for further processing in the second DFS stage. 

\noindent\textbf{Clustering.} The second stage employs a K-nearest neighbors density peaks clustering approach inspired by DPC-KNN \cite{du2016study,jin2024chat} to identify $K_c$ representative frames from $Z$. For each token $z_l$, we first compute its local density $\sigma_l$ based on its $K$-nearest neighbors:
\begin{equation}
    \sigma_l = exp(- \frac{1}{K}\sum_{z_k \in KNN(z_l, Z)} ||z_k - z_l||^2),
\end{equation}\label{eq:density}
where KNN($z_l$, $Z$) returns the $K$-nearest neighbors of $z_l$ from $Z-\{z_l\}$\footnote{$Z-\{z_l\}$ denotes removing $z_l$ from the set $Z$.}. Then, we compute each token's distance index $\rho_l$ of $z_l$:
\begin{equation}
    \rho_l = \begin{cases}
        \underset{j:\sigma_j > \sigma_l}{\min} ||z_j - z_l||^2 & \textit{if  }   \exists \textit{ j } \textit{ s.t. } \sigma_j > \sigma_l, \\
        \phantom{j}\underset{j}{\max} \phantom{j} ||z_j - z_l||^2 & otherwise. \\ 
    \end{cases}
\end{equation}
%


\noindent{In} essence, $\rho_l$ represents the distance between the given token $z_l$ from other high-density tokens. We then use $\sigma_l \times \rho_l$ as the weighted density index for each $z_l$ and sort them in descending order. The top $K_c$ frames with the highest indices are selected as the most representative video moments related to the user instruction. We use the indices of these $K_c$ centers to extract the representations with a higher spatial resolution for each frame from the feature buffer containing ViT encodings. Finally, these representations are pooled to a desired number of tokens per frame denoted as $\hat{Z}$. This mechanism effectively ``rewinds'' through the video's latent space to identify key moments, inspiring ReWind's name.


\subsection{Large Language Model}
\label{ss:llm}

The input to the LLM is constructed by concatenating $M$ with the dense representations $\hat{Z}$, separated by a special token $\tau$: $<m_0, m_1, \dots, \tau, \hat{Z}>$. The role of $\tau$ is purely to separate the memory content with progressive temporal information from the DFS frames where the spatial information is prioritized. The video content is then combined with the text instruction and given in input to the LLM.

\subsection{Training}
\label{ss:training}
Instruction tuning has been a crucial training strategy for VLMs, especially for the QA tasks, as demonstrated from previous works \cite{jin2024chat,li2023llamavid}. Inspired by this, our training strategy is divided into two stages.

\noindent\textbf{Multimodal Pretraining Stage.} During the initial stage, we conduct standard multimodal alignment, keeping the network components except the perceiver frozen. This phase aims to empower our ReWind to effectively capture semantic visual information without any
compromise in the performance of the overall pipeline.
Specifically, it involves contrastive learning utilizing the SigLIP \cite{zhai2023sigmoid} loss between perceiver projections and caption encodings from BERT. 

\noindent\textbf{Instruction Tuning Stage.} The second training stage engages the memory module, the DFS, and the LLM (fine-tuned using LoRA). This phase employs the instruction-tuning strategy on multimodal instruction-tuning datasets, 
aiming to integrate all network components seamlessly for the VQA and temporal grounding tasks.

\section{Experiments}
\label{s:experiments}

\begin{table*}[!ht]
    \centering
    \resizebox{\textwidth}{!}{%
        \begin{tabular}[l]{l l l c c c c c c  c ccccc c ccccc}
             \toprule
             \multirow{2}{*}{\textbf{Model}}& \multirow{2}{*}{\textbf{Num Frames}} & \multirow{2}{*}{\textbf{Num Tokens}} & &\multicolumn{2}{c}{\textbf{Global VQA}} && \multicolumn{2}{c}{\textbf{Breakpoint VQA}} && \multicolumn{5}{c}{\textbf{Global Generation}} && \multicolumn{5}{c}{\textbf{Breakpoint Generation}}\\  \cmidrule{5-6} \cmidrule{8-9} \cmidrule{11-15} \cmidrule{17-21}
             & & & & Accuracy & Score && Accuracy & Score && \textbf{CI} & \textbf{DO} & \textbf{CU} & \textbf{TU} & \textbf{CO} && \textbf{CI} & \textbf{DO} & \textbf{CU} & \textbf{TU} & \textbf{CO}\\ \midrule
             Video LLaMA \cite{zhang2023videollama} & 32 & 32 & & 51.4 & 3.10 && 38.2 & 2.31 && 3.30 & 2.53 & 3.28 & 2.77 & 3.42 && 2.42 & 2.85 & 2.87 & 2.00 & 2.87\\
             Video-ChatGPT \cite{maaz2024ChatGPT} & 100 & 356 & & 44.2 & 2.71 && 49.8 & 2.71 && 2.48 & 2.78 & 3.03 & 2.48 & 2.99 && \underline{3.11} & \underline{3.32} & 3.29 & 2.62 & 3.29\\
             Video Chat \cite{2023videochat} & 32 & 3072 & & 61.0 & 3.34 && 48.3 & 2.43 && 3.26 & 3.20 & 3.38 & 2.97 & 3.47 && 2.96 & 3.09 & 3.24 & 2.46 & 3.22 \\
             MovieChat \cite{song2024moviechat} & 2048 & 8192 & & \underline{67.8} & \underline{3.81} && \underline{50.4} & \underline{2.96} && \underline{3.32} & \underline{3.28} & \underline{3.44} & \underline{3.06} & \underline{3.48} && 3.07 & 3.24 & \underline{3.31} & \underline{2.70} & \underline{3.45}\\
             \textbf{ReWind (Ours)} & 548* & 1184* & & \textbf{80.6} & \textbf{4.46} && \textbf{57.2} & \textbf{3.4} && \textbf{4.18} & \textbf{4.00} & \textbf{4.24} & \textbf{4.02} & \textbf{3.54} && \textbf{3.41} & \textbf{3.37} & \textbf{3.64} & \textbf{2.97} & \textbf{3.61} \\
             \bottomrule
        \end{tabular}%
        }
    \caption{Evaluation for long VQA on MovieChat-1K test set with GPT-3.5. The best result is in \textbf{bold}, and the second best is \underline{underlined}. Rewind uses a fixed frame rate of 1fps, so the number of frames in input varies based on the video length. '*' means the quantity is variable.}
    \label{tab:long_qa}
\end{table*}

\begin{table}[!ht]
    \centering
    \resizebox{\linewidth}{!}{%
        \begin{tabular}[l]{l c cccc}
             \toprule
             \multirow{2}{*}{\textbf{Model}} && \multicolumn{4}{c}{\textbf{Charades-STA}} \\
             &&  R@0.3 & R@0.5 & R@0.7 & mIoU \\ \midrule
             Video Chat \cite{2023videochat} && 9.0 & 3.3 & 1.3 & 6.5 \\
             Video LLaMA \cite{zhang2023videollama} && 10.4 & 3.8 & 0.9 & 7.1 \\
             Video-ChatGPT \cite{maaz2024ChatGPT} && 20.0 & 7.7 & 1.7 & 13.7 \\
             GroundingGPT \cite{li2024groundinggpt} && - & 29.6 & 11.9 & - \\
            TimeChat \cite{ren2024timechat}&& - & {\color{gray}32.2} & {\color{gray}{13.4}} & - \\
             VTimeLLM \cite{huang2024vtimellm} && \underline{51.0} & \underline{27.5} & \underline{11.4} & \underline{31.2} \\
            \textbf{ReWind (Ours)} && \textbf{59.0} & \textbf{41.6} & \textbf{20.53} & \textbf{39.3}\\
             \bottomrule
        \end{tabular}%
        }
    \caption{Temporal video grounding on Charades-STA. Best results are emphasized in \textbf{bold}, and second-bests are \underline{underlined}. We compare against works that use only video inputs. {\color{gray}De-emphasized results use transcribed speech in input.}}
    \label{tab:temporal_grounding}
\end{table}

\subsection{Experimental Setup and Datasets}
\noindent{\textbf{Model Settings.}} ReWind's architecture is built upon the EVA-02 vision encoder (ViT-G/14) \cite{fang2023eva} and a 7B-parameter LLaMA-2 LLM \cite{touvron2023llama}. The perceiver block, illustrated in Fig. \ref{fig:arch} (b), consists of 8 sequential layers. Additionally, we utilize 32 queries for reading and perceiving information and two write queries to ensure efficient memory storage. The DFS mechanism selects 64 frames in the first selection phase and then refines this to 8  representative frames. These selected frames are then pooled into 32 tokens per frame before being integrated with the memory content.

\noindent\textbf{Training Setup and Data.}
We pretrain ReWind on 100K video-caption pairs randomly selected from the WebVid2.5M \cite{Bain21@webvid} and Panda70M \cite{chen2024panda} datasets. This stage involves 10K steps with a batch size of 64, using the AdamW optimizer and cosine scheduling. The learning rate is set to 1e-4 with 500 warmup steps. For instruction tuning, we combine multimodal instruction data from VideoChatGPT \cite{maaz2024ChatGPT} with the same 100{,}000 video-caption pairs used in the pretraining stage. All frames are resized to 224$\times$224 pixels. During this stage, ReWind is trained for 100{,}000 steps with a batch size of 64, a learning rate of 5e-5, and 2{,}000 warmup steps, using the same optimizer and scheduler as in pretraining. We utilize LoRA for the LLM with a rank of 64 and alpha of 32. For temporal grounding tasks, ReWind undergoes additional fine-tuned on DiDemo \cite{anne2017localizing} and ActivityNet \cite{caba2015activitynet} datasets with manually annotated QA pairs with temporal boundaries for an extra 15K steps using the same optimizer and learning rate. Remarkably, our model can obtain great results while being trained on only 8$\times$V100 GPUs. Further details regarding the data and training setup can be found in the supplementary material.

\subsection{Datasets and Evaluation}
\noindent \textbf{Long Video.}
We evaluate ReWind's performance on two tasks: VQA and temporal grounding. For VQA, we use the MovieChat-1K test set\cite{2023videochat}, with a video average length of 9.13 minutes. We assess VQA performance using three metrics: accuracy, score, and generation quality, determined by comparing the generated answer to the ground truth (GT) using GPT-3.5. Accuracy measures the exact matches between answers and GT, while the score measures their proximity in meaning with a score from 0 to 5. Generation quality is evaluated using the protocol proposed in \cite{2023videochat} based on five metrics: correctness of information (CI), detailed orientation (DO), contextual understanding (CU), temporal understanding (TU), and consistency (CO). Each metric is assigned a score from 0 to 5 by GPT-3.5 by comparing the generated answer and the GT. For temporal grounding, we use Charades-STA \cite{gao2017tall}. We measure recall at various thresholds (30-70\%) and mean IoU (mIoU) to compare the predicted time intervals with the GT.

\noindent\textbf{Short Video.} We evaluate ReWind's performance on short-video benchmarks using the VideoChatGPT dataset and generation quality evaluation protocol.

\begin{table}[!th]
    \centering
    \resizebox{\linewidth}{!}{%
        \begin{tabular}[l]{l  l l c cccccc}
             \toprule
             \textbf{Model} & \textbf{LLM} & \textbf{Backbone} & & \textbf{CI} & \textbf{DO} & \textbf{CU} & \textbf{TU} & \textbf{CO} & \textbf{AVG} \\ \midrule
             %
             %
             Video Chat \cite{2023videochat} & Vicuna-7B & ViT-G && 2.23 & 2.50 & 2.53 & 1.94 & 2.24 & 2.29 \\
             Video LLaMA \cite{zhang2023videollama} & Vicuna-7B & ViT-G && 1.96 & 2.18 & 2.16 & 1.82 & 1.79 & 1.98 \\
             Video-ChatGPT  \cite{maaz2024ChatGPT}& Vicuna-7B & ViT-L&& 2.40 & 2.52 & 2.62 & 1.98 & 2.37 & 2.38 \\
             LLaMA Adapter \cite{zhang2024llamaadapter} & LLaMA-7B  & ViT-L &&  2.03 & 2.32 & 2.30 & 1.98 & 2.15 & 2.16 \\
             Chat-UniVi \cite{jin2024chat} & Vicuna1.5-7B & ViT-L && 2.89 & 2.91 & \underline{3.46} & 2.39 & \textbf{2.81} & \underline{2.89}\\ 
             %
            VTimeLLM \cite{huang2024vtimellm}  & Vicuna1.5-7B & ViT-L && 2.78 & \textbf{3.10} & 3.40 & \underline{2.49} & 2.47 & 2.85 \\
             MovieChat \cite{song2024moviechat} & LLaMA2-7B & ViT-G && 2.76 & 2.93 & 3.01 & 2.24 & 2.42  & 2.67\\
             %
             LLaMA-VID \cite{li2023llamavid} & Vicuna-7B & ViT-G && \textbf{2.96} & \underline{3.00} & \textbf{3.53} & 2.46 & 2.51 & \underline{2.89}\\
             
              
           
             \textbf{ReWind (Ours)} & LLaMA2-7B & ViT-G && \underline{2.91} & 2.85 & 3.42 & \textbf{2.71} & \underline{2.68} & \textbf{2.91}\\
             \bottomrule
        \end{tabular}%
        }
    \caption{Evaluation for short VQA on VideoChatGPT test set with GPT-3.5. The best result in \textbf{bold}, and the second best \underline{underlined}.}
    \label{tab:short_qa}
\end{table}

\subsection{Results on Long Videos}
\noindent\textbf{VQA.}  The MovieChat-1K dataset is a challenging long-video benchmark with an average video length of 9.13 minutes. It contains 1,000 videos, each with multiple open-ended questions in two settings: global and breakpoint. The global setting requires processing the entire video and answering questions about its content, while the breakpoint mode involves processing the video up to a specific timestamp and answering questions about the event at that point. Table \ref{tab:long_qa} presents the results for both settings on the test set, showcasing ReWind's performance on generation quality and accuracy-score metrics. The analysis reveals that ReWind significantly outperforms previous approaches across all metrics, particularly surpassing MovieChat \cite{song2024moviechat}, specifically designed for long videos. Notably, ReWind achieves these superior results while utilizing approximately 1/8 of the tokens and 1/4 of the frames required by the prior best model. This demonstrates ReWind's ability to effectively model temporal relationships over extended sequences and its efficiency in encoding information with a minimal number of tokens.

\noindent\textbf{Temporal Grounding.} Charades-STA \cite{gao2017tall} test set contains manually annotated QA pairs with temporal boundaries, providing a challenging testbed for assessing a model's understanding of event dynamics. We benchmark ReWind against existing VLM approaches and report results in Table \ref{tab:temporal_grounding}. Notably,  ReWind significantly outperforms all previous models that rely solely on video input across all metrics. This highlights ReWind's exceptional ability to accurately track and interpret the temporal progression of events.

\subsection{Results on Short Videos}

\noindent To further assess ReWind's capabilities, we evaluate its performance on the VideoChatGPT QA test set, which features open-ended questions with more detailed answers. Utilizing the generation evaluation protocol, the results are presented in Table \ref{tab:short_qa}. ReWind achieves a higher overall average score (AVG) than all previous short and long-term methods, demonstrating strong performance even in short videos. Notably, ReWind excels in the temporal understanding (TU) metric, confirming its superior ability to capture and comprehend temporal information. More experiments can be found on the supplementary material.


\section{Ablation}
\label{s:ablation}

\noindent\textbf{Core Mechanisms.} Table \ref{tab:ablation_mem_dfs} presents an ablation of ReWind's core components --- the memory module, and the DFS mechanism --- on long videos. We establish a baseline model that uses 64 uniformly sampled frames and incorporates the perceiver block as an adapter layer, with each frame encoded using 32 tokens.  We then progressively incorporate the memory and DFS to complete ReWind's architecture. Note that when we add the components, the video is processed at 1 fps, and each frame is encoded with 2 tokens to align with our design. The results demonstrate that memory and DFS significantly contribute to ReWind's performance on long videos. To assess the effectiveness of these components on shorter videos, we conduct a similar ablation using the VideoChatGPT dataset, which consists of short videos, and report the findings in Table~\ref{tab:ablation_mem_dfs_chatgpt}.  Notably, combining memory and DFS leads to substantial improvements over the baseline, even when applied to short videos.

\begin{table}[]
    \centering
    \resizebox{\linewidth}{!}{%
        \begin{tabular}[l]{l  c c c c c c}
             \toprule
             \multirow{2}{*}{\textbf{Model}}& &\multicolumn{2}{c}{\textbf{Global}} && \multicolumn{2}{c}{\textbf{Breakpoint}} \\  \cmidrule{3-4} \cmidrule{6-7}
             & & Accuracy & Score && Accuracy & Score \\ \midrule
             Baseline & & 61.5 & 3.21 && 49.1 & 2.62 \\
             Mem & & 76.8 & 4.21 && 52.1 & 3.11 \\
             Mem+DFS & & \textbf{80.6} & \textbf{4.46} && \textbf{57.2} & \textbf{3.40} \\
             \bottomrule
        \end{tabular}%
        }
    \caption{Ablation study on how the memory mechanism (Mem) and the DFS affect the performance of Rewind on MovieChat-1K.}
    \label{tab:ablation_mem_dfs}
\end{table}

\begin{table}[]
    \centering
    \resizebox{\linewidth}{!}{%
        \begin{tabular}[l]{l  c c c  c c c c}
             \toprule
             \textbf{Model} & & \textbf{CI} & \textbf{DO} & \textbf{CU} & \textbf{TU} & \textbf{CO} & \textbf{AVG} \\ \midrule
             Baseline & & 2.54 & 2.72 & 3.27 & 2.46 & 2.60 & 2.72\\
             Mem & & 2.76 & 2.56 & 3.13 & 2.58 & 2.62 & 2.73\\
             Mem+DFS & & \textbf{2.91} & \textbf{2.85} & \textbf{3.42} & \textbf{2.71} & \textbf{2.68} & \textbf{2.91}\\
             \bottomrule
        \end{tabular}%
        }
    \caption{Ablation study on how the memory mechanism (Mem) and the DFS affect the performance of Rewind on VideoChatGPT.}
    \label{tab:ablation_mem_dfs_chatgpt}
\end{table}

\noindent\textbf{Perceiver.} In our architectural design, the perceiver layer is conditioned on the text and past information through reading queries. We validate the effects these elements have on the perceiver in Table \ref{tab:ablation_perceiver}. Particularly, we start with ReWind without DFS and experiment with different conditions.

\begin{table}[!ht]
    \centering
    \resizebox{\linewidth}{!}{%
        \begin{tabular}[l]{c  c c c c c c}
             \toprule
             \multirow{2}{*}{\textbf{Text}}& \multirow{2}{*}{\textbf{Read}} &\multicolumn{2}{c}{\textbf{Global}} && \multicolumn{2}{c}{\textbf{Breakpoint}} \\  \cmidrule{3-4} \cmidrule{6-7}
             & & Accuracy & Score && Accuracy & Score \\ \midrule
             \xmark & \checkmark & 74.7 & 4.06 && 51.2 & 3.09 \\
             \checkmark & \xmark & 69.1 & 3.76 && 48.7 & 2.81 \\
             \checkmark & \checkmark & \textbf{76.8} & \textbf{4.21} && \textbf{52.1} & \textbf{3.11} \\
             \bottomrule
        \end{tabular}%
        }
    \caption{Ablation study on conditioning the perceiver with text and read information (past content).}
    \label{tab:ablation_perceiver}
\end{table}

\begin{figure}[!t]
\includegraphics[width=\linewidth]{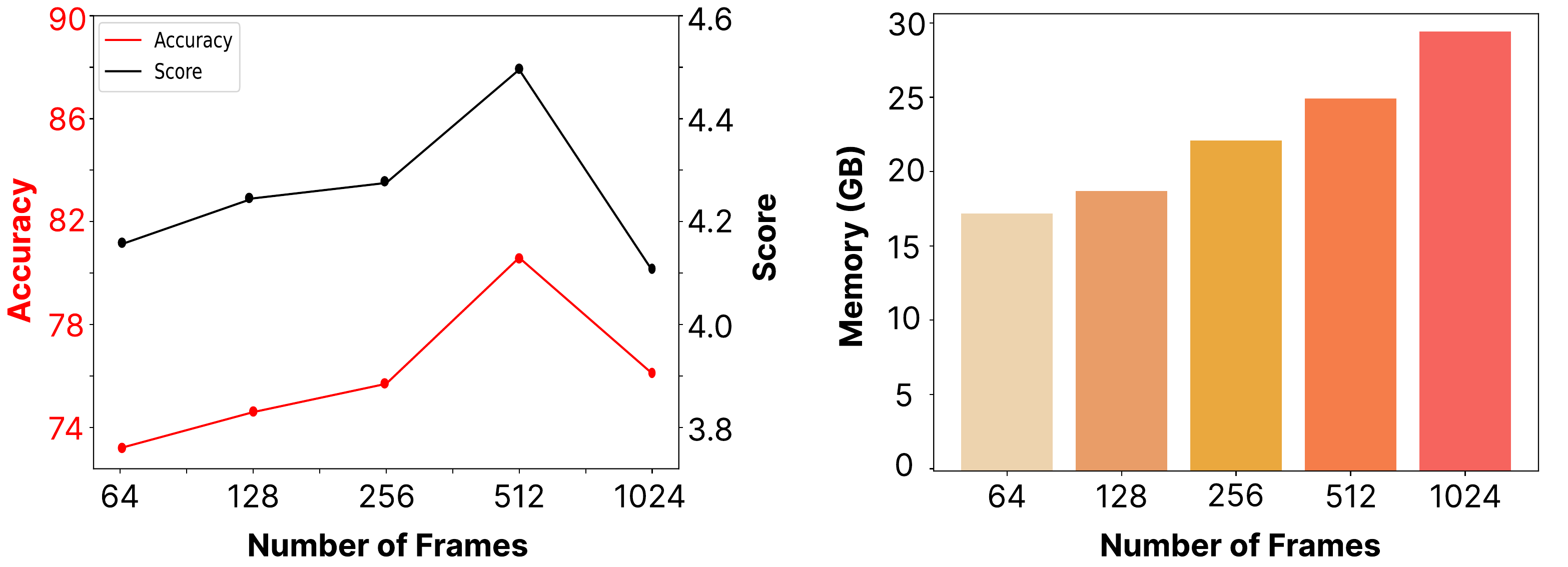}
 \caption{Ablation study on ReWind's performance and memory requirements in MovieChat-1K test set for different numbers of input frames, ranging from 64 to 1024,  and 16-bit precision.}
 \label{fig:num_frames}
\end{figure}

\noindent\textbf{Number of Frames vs. Performance and Memory.} Figure \ref{fig:num_frames} illustrates the impact of varying the number of input frames (ranging from 64 to 1024) on ReWind's performance and GPU memory requirements. ReWind's performance improves as the number of frames increases, reaching an optimal point at 512 frames (approximately 1 fps sampling). Beyond this point, performance declines when using 1024 frames (around 2 fps sampling). This decline is likely due to the deviation from ReWind's training regime and the introduction of high redundancy in token representations.

Memory consumption, measured using 16-bit precision, peaks at 29GB for 1024 frames. Notably, ReWind can process a ~10-minute video with less than 25GB of memory, making it compatible with standard end-user GPUs. Additionally, peak memory consumption is influenced by the choice of the LLM, and the number of input tokens. Utilizing different LLM quantizations (e.g., 8-bit) can substantially reduce memory requirements.

\begin{figure*}[!ht]
\includegraphics[width=\linewidth]{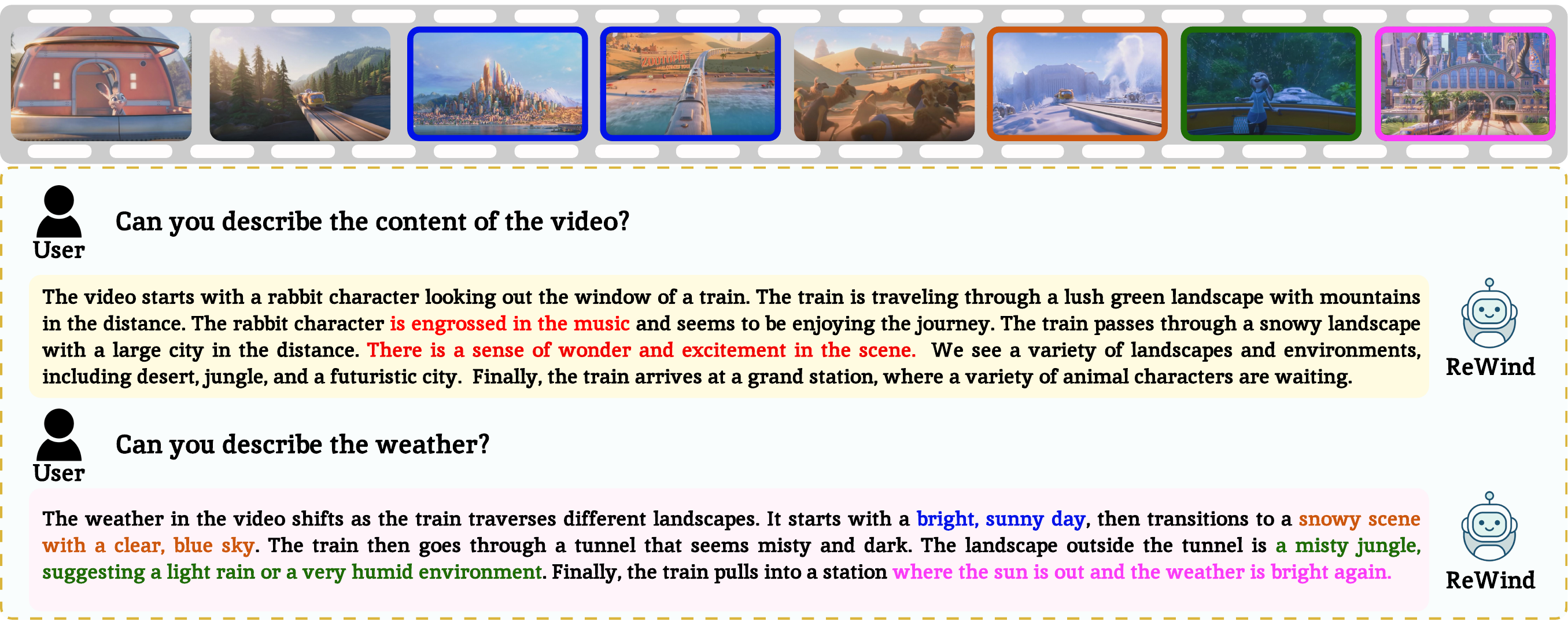}
 \caption{Qualitative result on VQA. We input ReWind with the illustrated video of +4 minutes and make two types of questions regarding the video content. On the first answer, we showcase ReWind's ability to understand the extended context and at the same time highlight in \textit{{\color{red}{red}}} the hallucination produced by it. In the second scenario, we highlight ReWind's ability to focus on different aspects of the video by matching some of the frames selected from DFS for the given scenario and the corresponding details on the generated answer.}
 \label{fig:study_case}
\end{figure*}

\noindent\textbf{Hyperparameters.} We ablate on the hyperparameters of ReWind to assess their impact. Initially, we vary the number of tokens per frame stored in memory without DFS to clearly understand its individual effect. The results are presented in Table \ref{tab:ablation_mtpf}. Furthermore, in Table \ref{tab:ablation_nfs}, we investigate DFS-specific hyperparameters: the number of selected tokens during the instruction-selection stage ($L$) and the number of final selected frames ($K_c$).

\noindent\textbf{DFS vs. Uniform Sampling.} Finally, we ablate the benefits of having DFS to uniform sampling for long videos. In both scenarios, the number of selected frames is 8. The outcomes of this comparison are detailed in Table \ref{tab:ablation_dfs}.

\begin{table}[!ht]
    \centering
    \resizebox{\linewidth}{!}{%
        \begin{tabular}[l]{l  c c c c c c}
             \toprule
             \multirow{2}{*}{\textbf{TPF}}& &\multicolumn{2}{c}{\textbf{Global}} && \multicolumn{2}{c}{\textbf{Breakpoint}} \\  \cmidrule{3-4} \cmidrule{6-7}
             & & Accuracy & Score && Accuracy & Score \\ \midrule
             1 & & 71.1 & 3.91 && 49.0 &2.87 \\
             2 & & 76.8 & 4.21 && 52.1 & 3.11 \\
             3 & & 79.2 & 4.34 && 54.7 & 3.34 \\
             4 & & \textbf{79.7} & \textbf{4.41} && \textbf{54.9} & \textbf{3.41} \\
             \bottomrule
        \end{tabular}%
        }
    \caption{Ablation study on the numbers of tokens per frame (TPF) stored in memory (write queries) on MovieChat-1K test set.}
    \label{tab:ablation_mtpf}
\end{table}

\begin{table}[]
    \centering
    \resizebox{\linewidth}{!}{%
        \begin{tabular}[l]{c c c c c c c c}
             \toprule
             \multirow{2}{*}{\textbf{$L$}}& \multirow{2}{*}{\textbf{$K_c$}} &&\multicolumn{2}{c}{\textbf{Global}} && \multicolumn{2}{c}{\textbf{Breakpoint}} \\  \cmidrule{4-5} \cmidrule{7-8}
             & & & Accuracy & Score && Accuracy & Score \\ \midrule
             16 & 8 & & 77.1 & 4.25 && 39.8 & 2.7 \\
             32 & 8 && 78.1 & 4.32 && 48.6 & 2.9 \\
             64 & 8 && \textbf{80.6} & \textbf{4.46} && \textbf{57.2} & \textbf{3.4} \\
             128 & 8 && 80.1 & 4.45 && 56.1 & 3.4 \\ \midrule \midrule
             64 & 4 & & 77.9 & 4.30 && 52.1 & 3.11 \\
             64 & 8 & & 80.6 & 4.46 && \textbf{57.2} & \textbf{3.40} \\
             64 & 16 & & \textbf{81.5} & \textbf{4.52} && 55.2 & 3.18 \\
             \bottomrule
        \end{tabular}%
        }
    \caption{Ablation study on the hyperparameters of the DFS mechanism. We explore the effects of varying the number of frames selected in the instruction-based selection stage.}
    \label{tab:ablation_nfs}
\end{table}

\begin{table}[]
    \centering
    \resizebox{\linewidth}{!}{%
        \begin{tabular}[l]{l  c c c c c c}
             \toprule
             \multirow{2}{*}{\textbf{FS Strategy}}& &\multicolumn{2}{c}{\textbf{Global}} && \multicolumn{2}{c}{\textbf{Breakpoint}} \\  \cmidrule{3-4} \cmidrule{6-7}
             & & Accuracy & Score && Accuracy & Score \\ \midrule
             Uniform & & 77.5 & 4.29 && 54.1 & 3.31\\
             DFS & & \textbf{80.6} & \textbf{4.46} && \textbf{57.2} & \textbf{3.40} \\
             \bottomrule
        \end{tabular}%
        }
    \caption{Ablation on frame selection (FS) strategy on MovieChat-1K test set. Comparison between DFS and uniform sampling.}
    \label{tab:ablation_dfs}
\end{table}

\subsection{Qualitative Results}
Fig. \ref{fig:study_case} provides qualitative examples showcasing ReWind's ability to comprehend long videos while preserving fine-grained details. We pose two types of questions: (1) a comprehensive description of the entire video content, where ReWind captures the overall narrative and key events, and (2) a question about the changing weather throughout the video, testing ReWind's ability to track and recall information across different scenes.  We highlight the model's hallucinations in red. In (2), we highlight selected frames from DFS and their corresponding text using matching colors.

\section{Conclusions}
\label{s:conclusions}

This work introduces ReWind, a novel memory-based vision-language model that enables an efficient understanding of long videos while maintaining temporal fidelity. ReWind features a dynamic learnable memory module with an innovative read-perceive-write cycle for instruction-guided feature encoding and robust temporal representation construction. Additionally, we propose an adaptive frame selection mechanism guided by memory contents to identify instruction-relevant key moments, enriching memory representations with detailed spatial information.  Our evaluation demonstrates significant performance gains on various long-video benchmarks, including visual question answering and temporal grounding tasks. These results highlight ReWind's effectiveness in comprehensive video understanding and its potential for real-world applications requiring deep temporal reasoning over extended video content.

\section*{Acknowledgments} 
We gratefully acknowledge Sami Remes for his invaluable contributions to this work. His expertise and insights, particularly during his time with the video understanding team at Huawei Helsinki Research Center, were instrumental in designing and implementing the perceiver block. We extend our sincere thanks for his dedication and support.
{\small
\bibliographystyle{ieee_fullname}
\bibliography{literature}
}



\end{document}


\maketitle

\section{Additional Short-Video VQA Experiments}
In Table~\ref{tab:short_vqa}, we report zero-shot results on MSRVTT-QA and MSVD-QA of ReWind. Our approach, despite being specifically designed for long videos, outperforms previous short-term video approaches and has competitive performance compared to the current SOTA. Specifically, we obtain SOTA results on \textit{Score} for both datasets. 

\begin{table}[!t]
    \centering
    \resizebox{\linewidth}{!}{%
        \begin{tabular}[l]{l  c c c c c c c c c}
             \toprule
             \multirow{2}{*}{\textbf{Model}} & \multirow{2}{*}{\textbf{LLM}} &  \multirow{2}{*}{\textbf{Backbone}} && \multicolumn{2}{c}{\textbf{MSRVTT-QA}} && \multicolumn{2}{c}{\textbf{MSVD-QA}} &\\  \cmidrule{5-6} \cmidrule{8-9}
             & & & & Accuracy & Score && Accuracy & Score &\\ \midrule
             \textit{Short-Video Approaches} & & & &  &  &&  &  & \\
             Video Chat \cite{2023videochat} & Vicuna-7B & ViT-G && 45.0 & 2.5 && 56.3 & 2.8 & \\
             Video LLaMA \cite{zhang2023videollama} & Vicuna-7B & ViT-G && 29.6 & 1.8 && 51.6 & 2.5 & \\
             Video-ChatGPT  \cite{maaz2024ChatGPT}& Vicuna-7B & ViT-L&& 49.3 & 2.8 && 64.9 & 3.3 &\\
             LLaMA Adapter \cite{zhang2024llamaadapter} & LLaMA-7B  & ViT-L && 43.8 & 2.7 && 54.9 & 3.1 &  \\
             Chat-UniVi \cite{jin2024chat} & Vicuna1.5-7B & ViT-L && 55.0 & 3.1 && 69.3 & \underline{3.7} & \\ \midrule \midrule
             \textit{Long-Video Approaches}& & & &  &  &&  &  & \\
             MA-LMM \cite{he2024ma} & Vicuna-7B & ViT-G && 48.5 & - && 60.6 & - &\\ 
            VTimeLLM \cite{huang2024vtimellm}  & Vicuna1.5-7B & ViT-L && - & -&& - & - & \\
             MovieChat \cite{song2024moviechat} & LLaMA2-7B & ViT-G && 52.7 & 2.6 && \textbf{75.2} & \textbf{3.8} &\\
             %
             LLaMA-VID \cite{li2023llamavid} & Vicuna-7B & ViT-G && \textbf{58.3} & \underline{3.3} && \underline{69.7} & \underline{3.7} &\\
             
              
           
             \textbf{ReWind (Ours)} & LLaMA2-7B & ViT-G & & \underline{56.9} & \textbf{3.4} && 69.3 & \textbf{3.8} & \\
             \bottomrule
        \end{tabular}%
        }
    \caption{Evaluation for short VQA on MSRVTT-QA, MSVD-QA, and VideoChatGPT test sets with GPT-3.5. The best result is highlighted in \textbf{bold}, and the second best is \underline{underlined}.}
    \label{tab:short_vqa}
\end{table}

\section{Pretraining Stage}
During the pretraining stage, we conduct standard multimodal alignment, to train the perceiver component to effectively capture 
semantic visual information. Specifically, as reported in the main manuscript, it consists of contrastive learning utilizing the SigLIP \cite{zhai2023sigmoid} loss between perceiver projections and caption encodings from BERT \cite{devlin2018bert}. Particularly, we use a pre-trained BERT from hugging face under the following repository name: '\textit{google-bert/bert-base uncased}'\footnote{https://huggingface.co/google-bert/bert-base-uncased}. In this stage, the ViT (EVA02-ViT-G/14) is kept frozen and the features are extracted from its penultimate layer as suggested in \cite{fang2023eva}.
As reported in the manuscript, the pertaining is done on 100K video-caption pairs randomly selected from the WebVid2.5M and Panda70M datasets. 

\noindent\textbf{WebVid2.5M.} WebVid2.5M is a large-scale video-text dataset comprising 2.5 million video-text pairs predominantly sourced from YouTube \cite{Bain21@webvid}. It encompasses diverse topics and genres, offering a representative sample of real-world video content. This dataset is instrumental for large-scale training of models for tasks such as video captioning, retrieval, question answering, and video-language pre-training. Notably, WebVid2.5M employs weakly labeled data, where text annotations are extracted from sources like titles and descriptions, presenting a realistic challenge for video-language models.

\noindent\textbf{Panda70M.} Panda-70M is a large-scale video-caption dataset. It consists of 70 million high-quality video-caption pairs, derived by splitting 3.8 million long videos from the HD-VILA-100M dataset into semantically coherent clips \cite{chen2024panda}.  Multiple cross-modality teacher models were employed to generate diverse captions for each clip, with a fine-grained retrieval model subsequently selecting the most relevant caption as the final annotation. Panda-70M is intended for large-scale training in tasks such as video captioning, video and text retrieval, and text-driven video generation.

\begin{table}[]
    \centering
    \resizebox{\linewidth}{!}{%
    \begin{tabular}{c|c|c|c}
        \toprule
             \textbf{Dataset} & \textbf{ \#Clips} & \textbf{Avg. Duration (sec.)} & \textbf{Avg. Text len} \\ \midrule
        WebVid2.5M \cite{Bain21@webvid}      & 2.5M            & 18   &  12 \\ 
        Panda-70M \cite{chen2024panda}       & 70.8M             & 8   &    13.2  \\ 
             \bottomrule
        \end{tabular}%
        }
    \caption{Comparison of WebVid2.5M and Panda-70M Datasets.}
    \label{tab:panda_vs_webvid}
\end{table}

Additionally, as mentioned in the manuscript, this stage involves 10K training steps with a batch size of 64, using the AdamW optimizer and cosine scheduling. The learning rate is set to 1e-4 with 500 warmup steps. The entire training process is done in half-precision on 8 GPUs. For each video, in this stage, we uniformly sample 64 frames and resize the frames to 224 $\times$ 224.

\section{Instruction Tuning}
\label{s:instruction}
The second training stage engages also the memory module, the DFS, and the LLM (LLaMA-2, from huggingface: '\textit{"meta-llama/Llama-2-7b-chat-hf"}\footnote{https://huggingface.co/meta-llama/Llama-2-7b-chat-hf} with LoRA. This phase employs the instruction-tuning strategy on on $\sim$100K samples from the VideoChatGPT \cite{2023videochat} dataset together with the previous 100K samples used for pertaining, aiming to integrate all network components seamlessly for the VQA. 

\noindent\textbf{VideoChatGPT.} The Video-ChatGPT dataset comprises approximately 100K video-instruction pairs designed to enhance multimodal conversational AI models \cite{2023videochat}. The annotations combine human-assisted efforts and semi-automatic methods. In the human-assisted stage, expert humans are used to enrich the details of the existing ActivityNet \cite{caba2015activitynet} dataset. On the other hand, during the semi-automatic stage, using advanced vision-language models like BLIP-v2 \cite{li2023blip} and GRIT enables scalable and high-quality labeling of spatial, temporal, and contextual video content. 

During this stage, as reported in the main paper ReWind is trained for 100K steps with a batch size of 64, a learning rate of 5e-5, and 2K warmup steps, using the AdamW optimizer. LoRa is configured with a rank of 64 and an alpha of 32 and quantized in 4-bit during training. The training is done on 8 V100 GPUs. Additionally, the frames are selected with a constant frame rate of 1fps and are not bound to a fixed number of frames.

For the temporal grounding task, ReWind is further trained with instruction tuning on DiDemo and ActivityNet datasets for an extra 15K steps. During this training stage, we use 500 warmup steps and a batch size 64.

\noindent\textbf{DiDeMo.} The DiDeMo \cite{anne2017localizing} dataset is a large and diverse benchmark designed for temporally localizing events in videos based on natural language descriptions. It consists of videos collected from Flickr, each trimmed to a maximum duration of 30 seconds. These videos are segmented into 5-second intervals to simplify the annotation process. The dataset includes a total of 26,892 moments, with each moment possibly linked to multiple textual descriptions, offering detailed accounts that often specify camera movements, temporal transitions, and activities.

\noindent\textbf{ActivityNet.} The ActivityNet \cite{caba2015activitynet} dataset is a large-scale video benchmark for temporal action localization, captioning, and VQA. It consists of approximately 20,000 untrimmed videos sourced from YouTube, covering 200 different activity classes. Each video typically contains an average of 1.41 annotated activities, with temporal boundaries provided for precise action localization. 

\section{Evaluation Datasets.}
We evaluate our model on three datasets: MovieChat1K, Charades-STA, and VideoChatGPT dataset. Here, we provide information regarding MovieChat1K and Charades-STA. For VideoChatGPT, please refer to Section \ref{s:instruction}.

\noindent\textbf{MovieChat1K.}
The MovieChat1K \cite{song2024moviechat} dataset is a benchmark for evaluating video understanding through long-form video question answering. It comprises 1K long videos (9.13 minutes on average) annotated with 14K manually created questions and answers. The dataset is tailored to assess models' capabilities in processing extensive temporal contexts and understanding complex visual narratives in videos. It supports two VQA modalities: global and breakpoint. The global setting requires processing the entire video and answering questions about its content, while the breakpoint mode involves processing the video up to a specific timestamp and answering questions about the event at that point. Each video has one dense description, 3 global questions, and 10 breakpoint questions with timestamps.

\noindent\textbf{Charades-STA.} The Charades-STA dataset is designed for temporal activity localization in videos based on natural language queries. It is derived from the Charades \cite{gao2017tall} dataset and includes temporal annotations linking video clips to textual descriptions. Each query is paired with a specific start and end time within a video, allowing models to learn to locate activities matching the query. The dataset consists of 12,408 video-sentence pairs for training and 3,720 pairs for testing, with videos featuring everyday indoor activities (157 activity categories).

\section{Additional Qualitative Esamples}
\begin{figure*}
\includegraphics[width=\textwidth]{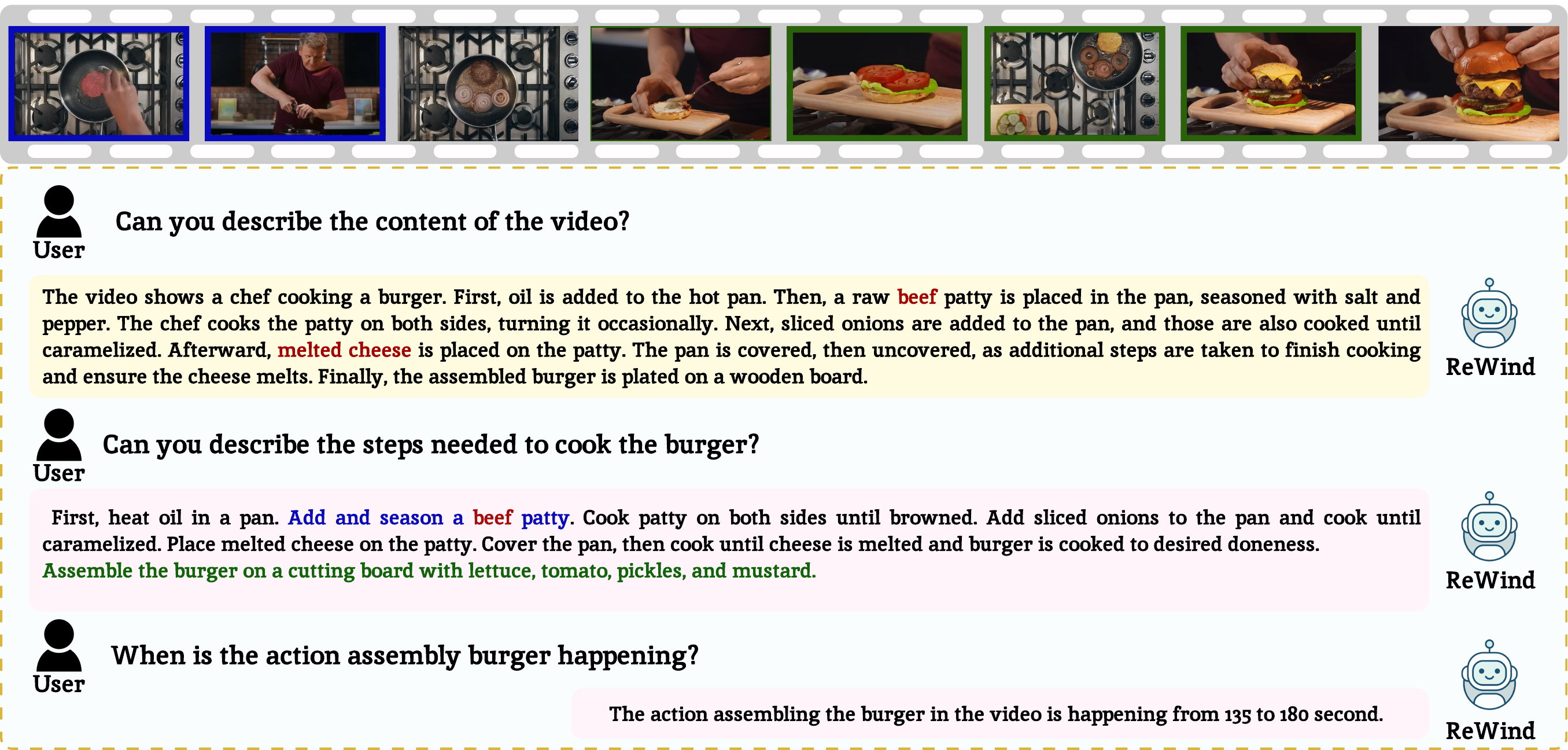}
 \caption{Qualitative evaluation using a cooking video over 3 minutes long, ReWind successfully described both the overall video content and the specific steps performed, demonstrating accurate temporal localization of events. To highlight potential areas of improvement, details that may be hallucinated based on language priors are indicated in red text.  Highlighted frames were selected using the DFS mechanism, with text color corresponding to the frame to easily connect descriptive details to their visual counterparts in the video.
}
 \label{fig:scase2}
\end{figure*}

\begin{figure*}[!b]
\includegraphics[width=\textwidth]{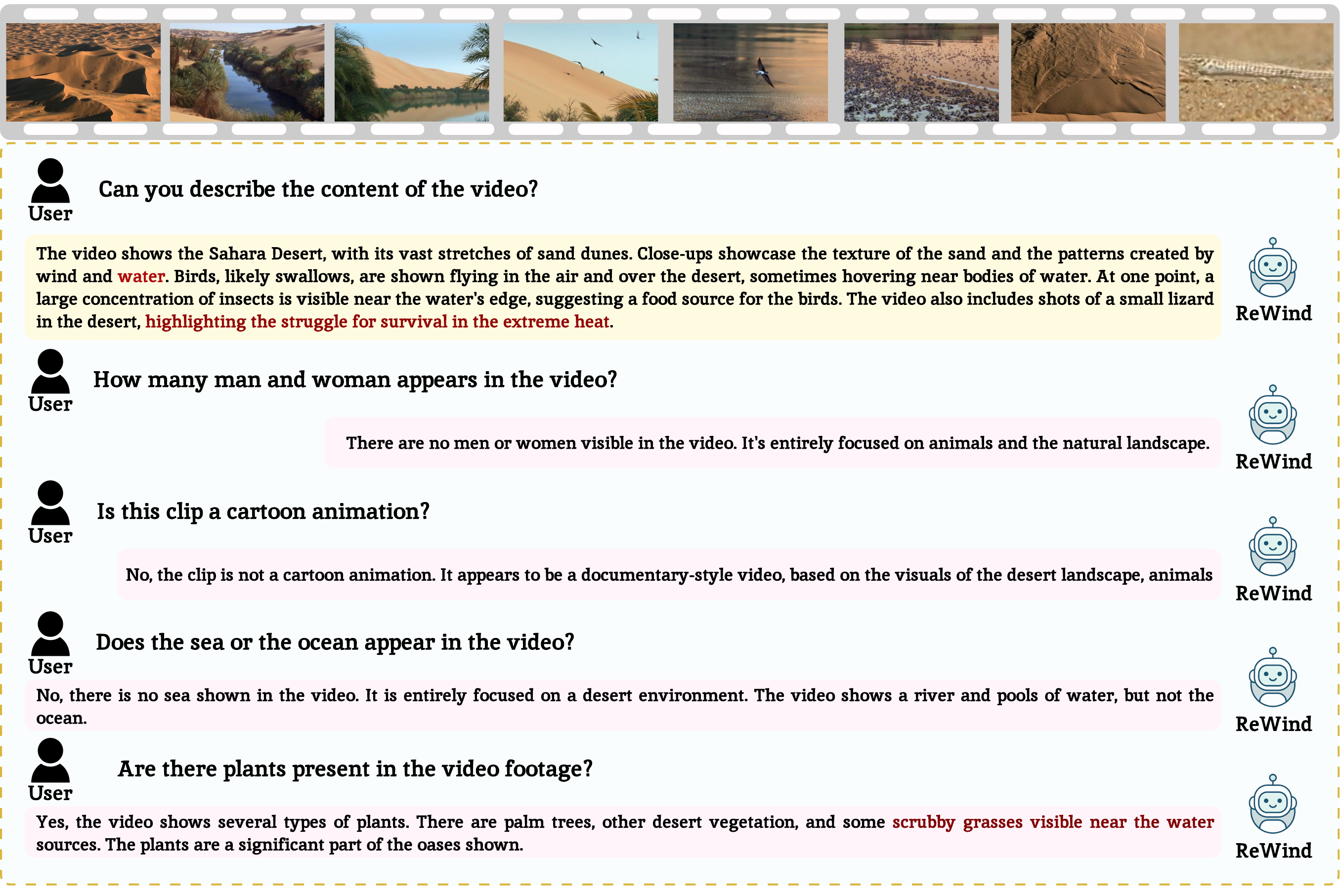}
 \caption{Qualitative evaluation on an +8 minutes video from MovieChat1K test set. ReWind is able to process this long video and answer questions of different natures, including a detailed description. Details hallucinated based on language priors are indicated in red text}
 \label{fig:scase3}
\end{figure*}

\begin{figure*}[!ht]
\includegraphics[width=\textwidth]{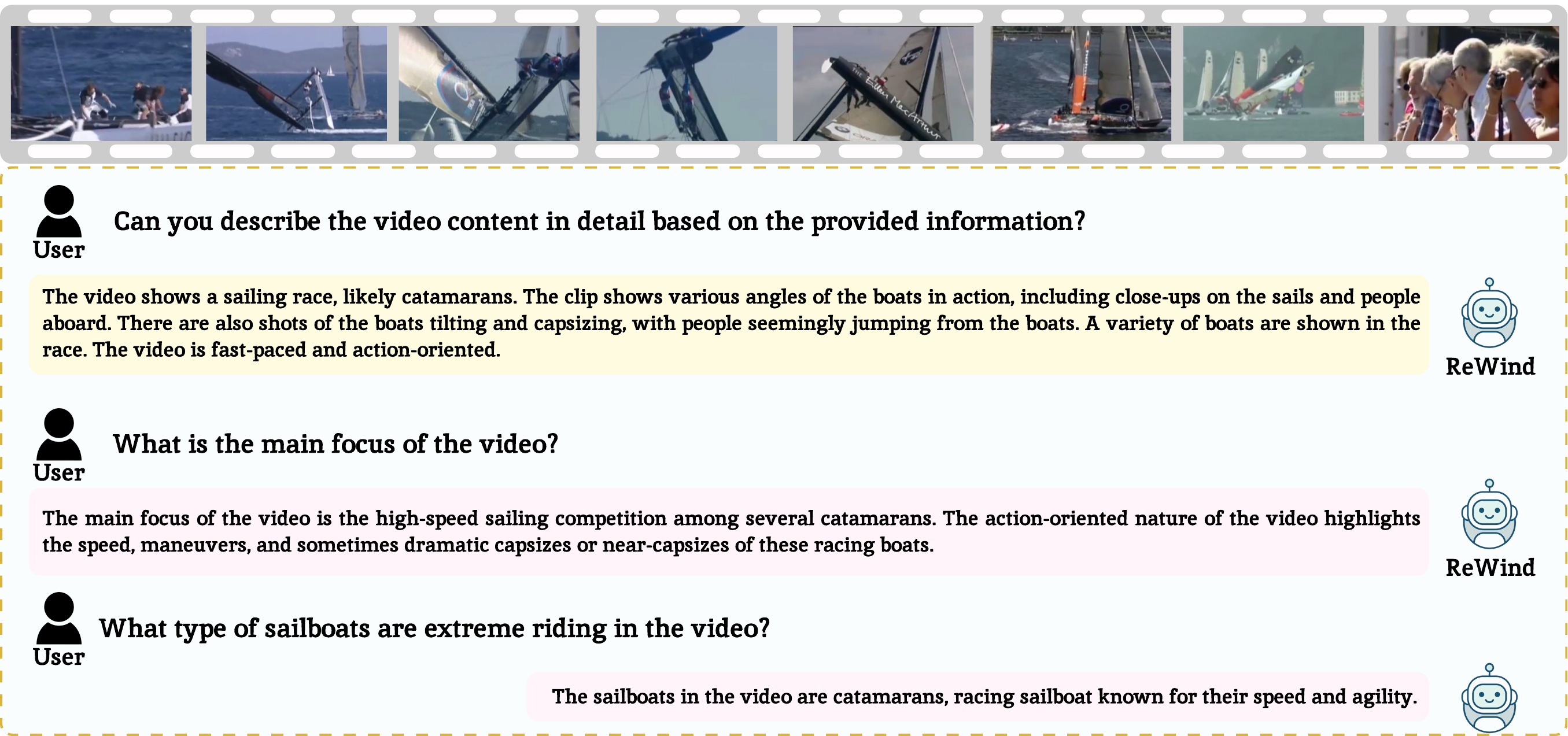}
 \caption{Qualitative example on a +3 minutes video from VideoChatGPT dataset. ReWind is able to correctly extract necessary information and answer to questions.}
 \label{fig:scase4}
\end{figure*}

Figures \ref{fig:scase2}, \ref{fig:scase3}, and \ref{fig:scase4} present further qualitative evaluations of ReWind, complementing the results within the manuscript. These figures illustrate three distinct cases: First, Figure \ref{fig:scase2} demonstrates ReWind's capabilities in a cooking scenario, showcasing its ability to describe video content, temporally localize actions, and highlight frames selected via DFS. Second, Figure \ref{fig:scase3}  evaluates ReWind's VQA proficiency in long videos using an 8-minute video from the MovieChat1K test set. Finally, Figure \ref{fig:scase4} assesses ReWind's VQA performance on a video from the VideoChatGPT validation set.

{\small
\bibliographystyle{ieee_fullname}
\bibliography{literature}
}